\documentclass{article}

\PassOptionsToPackage{compress}{natbib}
\usepackage[preprint]{neurips_2023}

\usepackage[utf8]{inputenc} 
\usepackage[T1]{fontenc}    
\usepackage{hyperref}       
\usepackage{url}            
\usepackage{booktabs}       
\usepackage{amsfonts}       
\usepackage{nicefrac}       
\usepackage{microtype}      
\usepackage{xcolor}         
\usepackage{amsmath, amssymb} 
\usepackage{caption}
\usepackage{subcaption}
\usepackage{graphicx}
\usepackage{floatrow, wrapfig} 
\usepackage{hhline}

\title{On the Dynamics of Learning\\Time-Aware Behavior with RNNs}

\author{%
Peter DelMastro$^{1*}$ \quad Rushiv Arora$^{2*}$ \quad Edward Rietman$^2$ \quad Hava T Siegelmann$^2$ \\
$^1$Department of Mathematics \& Statistics \\ 
$^2$College of Information and Computer Sciences \\
University of Massachusetts Amherst \\
\texttt{\{pdelmastro,rrarora,erietman,hava\}@umass.edu}
}

\begin{document}

\maketitle

\begin{abstract}
Recurrent Neural Networks (RNNs) have shown great success in modeling time-dependent patterns, but there is limited research on their learned representations of latent temporal features and the emergence of these representations during training. To address this gap, we use timed automata (TA) to introduce a family of supervised learning tasks modeling behavior dependent on hidden temporal variables whose complexity is directly controllable. Building upon past studies from the perspective of dynamical systems, we train RNNs to emulate \emph{temporal flipflops}, a new collection of TA that emphasizes the need for time-awareness over long-term memory. We find that these RNNs learn in phases --- they quickly perfect any time-independent behavior, but they initially struggle to discover the hidden time-dependent features. In the case of periodic ``time-of-day’’ aware automata, we show that the RNNs learn to switch between periodic orbits that encode time modulo the period of the transition rules. We subsequently apply fixed point stability analysis to monitor changes in the RNN dynamics \emph{during training}, and we observe that the learning phases are separated by a bifurcation from which the periodic behavior emerges. In this way, we demonstrate how dynamical systems theory can provide insights into not only the learned representations of these models, but also the dynamics of the learning process itself. We argue that this style of analysis may provide insights into the training pathologies of recurrent architectures in contexts outside of time-awareness.
\end{abstract}

\section{Introduction}

Recurrent neural networks (RNNs) \citep{elman90}, long-short term-memory networks \citep{hochreiter97}, and gated recurrent networks \citep{chung14} are some of the most widely used machine learning models for learning temporal relationships. Their ability to store and manipulate external inputs over time have made them popular in sequence related tasks such as time series prediction, language translation, or control.

Despite time-dependence being central to recurrent architectures, past literature has placed little emphasis on how artificial RNNs learn to utilize time itself in their computations \citep{Bi2020}. Such an ability to define and deduce temporal patterns is a fundamental skill required of artificial and biological agents. For instance, humans do not view each year as a sequence of 365 distinct days; rather, we decompose time into smaller repeating blocks based on some measure of periodicity (weeks) or counting (up to 7 days). Similarly, pet dogs often learn their owners daily and weekly schedules without explicitly being taught the concept of "hours" and "weeks". Architectures that can learn these modular representations of time will allow for smaller models with higher accuracy, faster learning, and increased generalization.

We ask if recurrent networks are able to be similarly temporally aware, and if so, how do they reveal the hidden dynamics. To address these questions, we use timed automata (TA) \citep{alur94} to introduce a new family of time-aware sequence processing tasks that give researchers direct control over the complexity of the time-awareness needed to solve them. These tasks are designed to allow scalability and customization which enable the testing of different forms of time-awareness.

Our work draws upon the long history of defining computational capabilities through automata. It extends past research on training neural networks to emulate automata behavior \citep{Pollack1991, Tino1998, zeng93, arai00, michalenko2019, oliva21, dan22} to include the time-dependent behavior described by timed automata introduced by \citet{alur94}.

In this study we focus on a type of timed automata that we call \textit{temporal flipflops} (TF). Time-independent flipflop automata were first studied in the context of RNNs by \citet{sussillo09}. They are characterized by symbols/actions which cause transitions that are independent of the previous states. Temporal flipflops retain this state-independence but their transitions also depend on latent temporal properties. This property allows us to \emph{isolate} the effects of temporal features on learning from more general long-term memory requirements.

Our analysis also extends existing research using tools from dynamical systems theory to analyze RNNs trained to emulate automata . In past studies on time-\textit{independent} automata, it was found that dynamics about stable fixed points encode the automata states, and input symbols induced state transitions by switching the networks' states between basins of attractions. \citet{sussillo13} discovered heteroclinic orbits connecting these fixed points, and \citet{ashwin20} used this behavior to construct continuous-time RNNs with behavior like finite-state machines. Spiking neural networks have also been modeled as timed automata \citep{demaria20}, but to our knowledge, such networks have never before been explicitly trained to emulate the behavior of time-dependent automata.

We analyze the dynamics of recurrent networks both \textit{during} and \textit{after} learning to shed light on the emergence of temporal representations. Post training we find that the networks learn reusable behaviors of time that significantly improve learning and generalization, with the entire sequence quantized into the smallest time period required to express the rhythms. In the case of periodic ``time-of-day'' aware automata we also observe distinct learning phases during training that coincide with a bifurcation in the RNNs' dynamics. In this way, our work is one of the first examples demonstrating how dynamical systems theory can provides insights into both the learned representations of these models and the dynamics of the learning process itself.

\section{Temporal Flipflop}

In this section we introduce temporal finite automata (TA), an approach to defining a wide class of time-aware computational capabilities. We start with a general overview of this formalism and then build the specific example of Temporal Flipflop---a state-independent TA.

\subsection{TA Formalism}

A \textit{temporal finite-state automaton} (TA) task is a time-series processing problem characterized by a tuple $(\Sigma, S, \Delta, s_0)$. Each instance of the task is a pair of discrete time functions $(u,y)$ where $u : \{1,...,T\} \to \Sigma$ is the input signal, which takes values in the finite set of symbols $\Sigma$; and $y : \{1,...,T\} \to S$ is the associated target output signal, which takes values in the finite set of automaton states $S$.

The function $y$ depends deterministically on $u$ through the time-dependent transition rule $\Delta : \Sigma \times S \times \mathbb{N} \to S$. We set $y(0) = s_0 \in S$ (the start state), and subsequent values of $y$ are generated according to the recurrence relation $y(t+1)= \Delta(y(t),u(t+1),t)$. This definition extends the standard deterministic finite automata (DFA) by making the transition rule to change with time, thus enabling for time-dependent state transitions.

Although many forms of time-dependence are possible under the TA framework, we elect to follow the clock-based formalism of timed automata \citep{alur94}. Our work focuses on transitions rules that depend on time through binary functions of the clock values. Specifically, we define an underlying temporal variable $\Theta : \{1, ..., T\} \to \{0,1\}$ computed from the clocks of the timed automaton, and the overall the transition rule $\Delta$ can be described by two time-independent transitions rules $\delta_0, \delta_1 : \Sigma\times S \to S$ that are applied when $\Theta = 0$ and $\Theta = 1$, respectively:
\begin{equation}
\Delta(c,s,t) = \delta_{\Theta(t)} = \begin{cases}
\delta_0(c,s) \ , & \Theta(t) = 0 \\ 
\delta_1(c,s) \ , & \Theta(t) = 1
\end{cases}
\end{equation}

\subsection{Types of Time-Dependence}
\label{sec:TF-Time-Dependence}

We consider two types of time-dependence for timed automata: periodic and relative timing.

\textbf{Periodic Timing:} We first construct TA that emulates “time of-day”-aware behavior. For these machines, time is divided into contiguous days of \emph{P} timesteps, and each day is further divided into two phases of duration \emph{P / 2}, called the day(light) and night phases. For instance, time $t \in \{ 1, \dots, P \}$ represents day 1, $t \in \{P + 1, P + 2, \dots , 2P\}$ represents day 2, etc. The temporal variable is defined to be a square wave --- $\Theta_\textrm{periodic}(t) = 0$ if $t \ (\textrm{mod} \ P) < P/2$ and $\Theta_\textrm{periodic}(t) = 1$ otherwise --- so the automaton uses the transition rules $\delta_0$ and $\delta_1$ during the day and night phases, respectively. 

\textbf{Relative Timing:} We also study a second TA that includes transitions which are dependent on the number of timesteps since an ‘event’ occurs, thereby enabling the automaton to count time. For these TA, we introduce a special “null” symbol $\phi$ that causes no change in state ($\Delta(\phi,s,t) = s$ for all $s \in S$ and $t$) and we equip these automata with a clock $c(t)$ that tracks the number of timesteps since the last \emph{non-null} symbol was received. The temporal variable $\Theta(t)$ indicates whether this clock exceeds a fixed threshold $\tau$: $\Theta_\textrm{relative}(t) = 0$ if $\phi$ last appeared \textit{at most} $\tau$ timesteps ago and $\Theta_\textrm{relative}(t) = 1$ otherwise. In this way, the TA responds to inputs by following either $\delta_0$ or $\delta_1$ based on the amount of time since it last saw a non-null symbol. To ensure that we have nearly equal probability of seeing both values of $\Theta(t)$, the TA is designed to receive non-null symbols with a probability $p$ that we arbitrarily choose.


\subsection{Flipflop Machines (State-\textit{Independent} TA)}

Given past research on the dynamics learned by RNNs trained to emulate the flipflop automata \citep{sussillo09,sussillo13}, the TA studied in this paper resembles the state-\emph{independent} behavior of these machines.

\begin{figure}[!t]
    \centering
    \begin{subfigure}{0.3\textwidth}
         \centering
         \includegraphics[width=\textwidth]{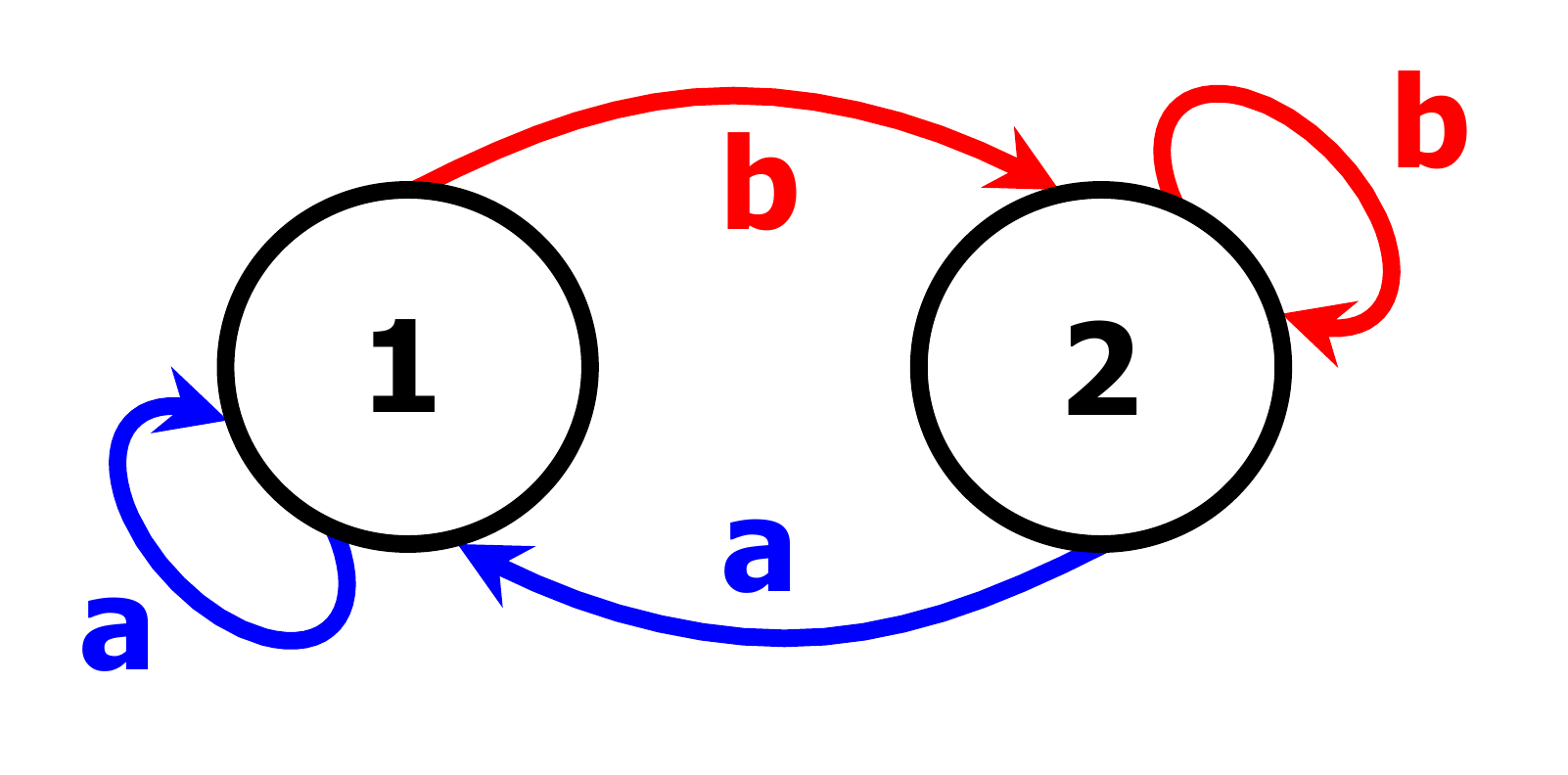}
         \caption{FF Transition Rule $\delta_0$}
         \label{fig:y equals x}
    \end{subfigure}
    \qquad
    \begin{subfigure}{0.3\textwidth}
         \centering
         \includegraphics[width=\textwidth]{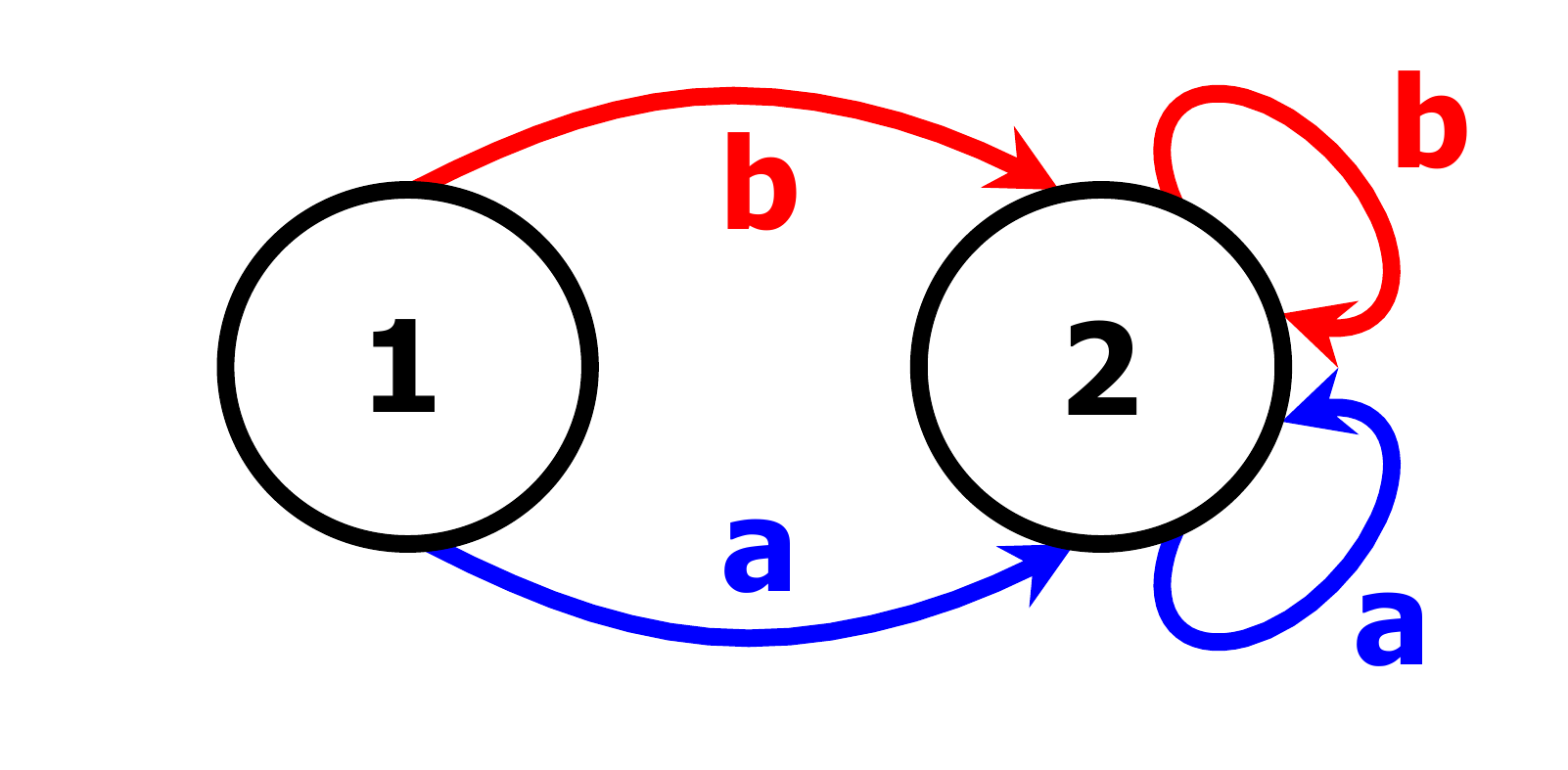}
         \caption{FF Transition Rule $\delta_1$}
         \label{fig:y equals x}
    \end{subfigure}
    \caption{\textbf{State Transition Diagrams for 2-State Flipflop TA.}  Each node in the graphs represent one of the states in $S = \{1,2\}$, and a directed edge \emph{(i,j)} with label $l \in \Sigma = \{\textit{\textcolor{blue}{a}, \textcolor{red}{b}}\}$ denotes a transition to state \emph{j} when symbol \emph{l} is received in state \emph{i}. Notice that transitions into State 2 are the same for both $\delta_0$ and $\delta_1$, so they are \textcolor{red}{time-\emph{independent}}.  Transitions into State 1 are not shared by $\delta_0$ and $\delta_1$; these are \textcolor{blue}{time-\emph{dependent}} transitions.}
    \label{fig:flipflop_TA}
\end{figure}

The (2-State) \emph{Temporal Flipflop} TA in Fig. \ref{fig:flipflop_TA} has states $S = \{1, 2\}$ and input symbols $\Sigma = \{a, b\}$. The transition rules $\delta_0$ and $\delta_1$ for this TA are quite simple: Symbol \emph{a} causes the automaton to transition to State 1 when $\Theta = 0$, and it causes transitions to State 2 when $\Theta = 1$. Symbol \emph{b} causes the automaton to transition to State 2, regardless of the value of $\Theta$. 

Two important considerations went into the design of the temporal flipflop. First, the task retains the state-independent transitions. This property reduces the long-term memory required to solve the task and hence attempts to minimize the vanishing gradient problem associated with learning long-term dependencies \citep{bengio94}.  Second, the TF has the capacity to exhibit both \textcolor{red}{time-\emph{independent}} and \textcolor{blue}{time-\emph{dependent}} behavior. Indeed, in Fig. \ref{fig:flipflop_TA}, \textcolor{red}{Symbol \emph{b}} always induces transitions to State 2, but transitions induced by \textcolor{blue}{Symbol \emph{a}} may lead to either States 1 or 2 depending on the value of the temporal variable $\Theta$. For this reason, training an RNN to emulate the temporal flipflop provides insight into the network’s abilities to efficiently learn these two types of behavior.




\section{Supervised Learning of Temporal Flipflop Behavior}

We now describe the supervised learning framework we use to train RNNs to emulate the temporal flipflop automata. Our approach focuses on \textit{hidden} temporal variables that the RNNs must learn to model internally in their hidden layers.

\textbf{TA Emulation by an RNN} \hspace{0.25cm}  For a TA $(\Sigma,S,\Delta,s_0)$, the sequence of symbols $u(1),u(2),\dots, u(T)$ will produce a sequence of states $y(1),y(2),\dots,y(T)$ according to the transition rule $y(t+1)=\Delta (u(t+1),y(t),t)$. A recurrent neural network will respond to the same input sequence with a hidden state sequence $h(1),\dots,h(T)$ generated according to the update rule $h(t)=F_h (u(t),h(t-1))$ and produces the associated output sequence $y_\textrm{out} (t)=F_y (h(t))$ where $F_h$ and $F_y$ are parameterizable functions that depend on the RNN architecture. The goal of training the network is to tune these functions such that $y_\textrm{out}(t)$ matches $y(t)$. 

\begin{figure}[t]
  \centering
  \includegraphics[width=0.9\textwidth]{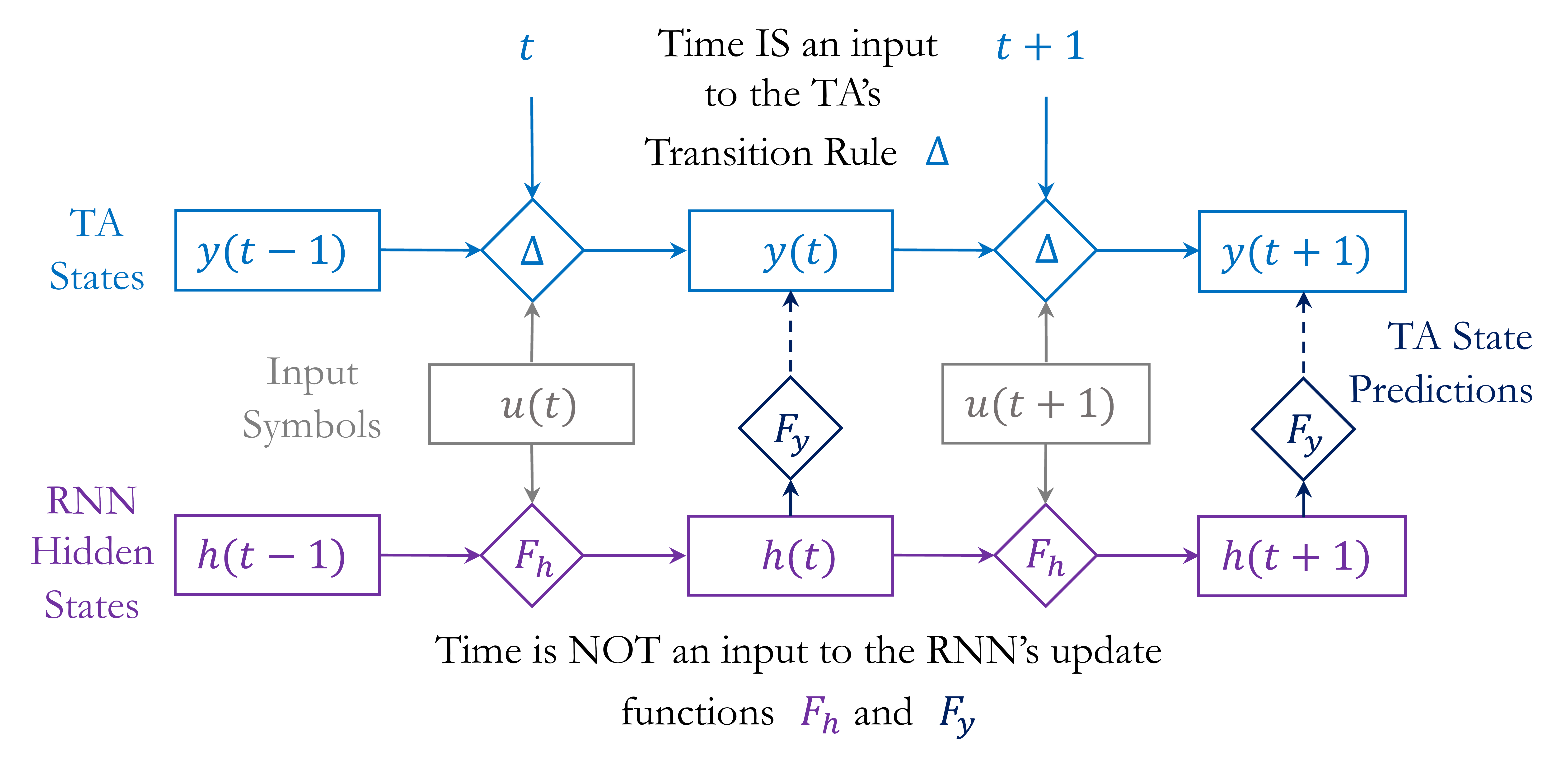}
  \caption{\textbf{Hidden time-dependence of the supervised TA tasks}. The network receives a sequence of symbols $u(t)$ as input which drives hidden state updates $h(t)=F_h (h(t-1),u(t))$. From these hidden states the RNN must be capable of computing the correct TA state $y(t)$ as its output $F_y (h(t))$. The key concept here is that the time-dependence of the TA transition rule $\Delta$ is hidden from the RNN. Whereas time is an explicit input to the TA, the RNN only updates based on input symbols and its past hidden state. The network must learn to represent the temporal information of the TA in the hidden state sequence.}
  \label{fig:SupervisedLearningDiagram}
\end{figure}

A \textbf{key concept} here is that the time-dependence of the TA transition rule $\Delta$ is hidden from the RNN. Whereas time is an explicit input to the TA, the RNN only updates based on input symbols and its past hidden state, as shown in Figure \ref{fig:SupervisedLearningDiagram}. The input sequence itself contains no information about the time-dependence of the automaton, so the network must learn to represent this temporal information through its hidden state sequence.

\textbf{Training} of the RNNs is performed using a supervised learning approach. We first generate a dataset of input-output examples $\mathcal{D}=\{(u^i,y^i)\}$ of the TA. The input sequences generated randomly with each symbol $u^i(t)$ drawn uniformly at random from the alphabet $\Sigma$ unless otherwise stated and the output sequences $y^i$ are computed from $u^i$ and the transition rule $\Delta$. We then train the RNN using stochastic gradient descent to learn this input-output mapping of sequences.

Because time is not an explicit input to the network in our learning framework, the time-dependence of the transition rule cannot be inferred directly from the input sequence nor the output sequence individually; instead, it only when these two sequences are considered together that the time-dependence becomes clear. In this way, TA tasks are characterized by \textit{temporal latent variables} that the network must discover and subsequently model through the training process.

\textbf{Evaluation Metrics} \hspace{0.25cm} The temporal flipflop automata involve state-\textit{independent} behavior because state transitions depend only on the input symbols and time of day. The two symbols can also be characterized as \textit{time-dependent} and \textit{time-independent}: Symbol $a$ causes transitions to State 1 or State 2 depending on time of day, whereas Symbol $b$ always leads to State 2. With this in mind, we define two metrics for evaluating RNNs trained to emulate the flipflop TA --- \textcolor{blue}{Time-dependent (TD)} and \textcolor{red}{Time-independent (TI) Accuracy}: the accuracy of the network’s predictions on timesteps when Symbol $a$ and Symbol $b$, respectively, was received as input. We continually compute the TD and TI accuracy of the network at every training iteration, and the resulting TI and TD learning curves are informative in monitoring the rate at which the networks learn time-dependent the behavior.


\section{Learning Periodic Time-Dependence}
\label{PeriodicFF}

We now present results collected when training RNNs to emulate the time-of-day aware temporal flipflop with $P=10$ timesteps per day. Here, we focus only on  Vanilla RNNs \citep{elman90} with a single fully-connected hidden layer of sizes $N_h=32$, but we observed similar results with both the GRU and LSTM. Refer to the Supplemental Material for further dataset and training details.

\subsection{Three Observed Phases of Learning}

\begin{figure}[t!]
    \floatbox[{\capbeside\thisfloatsetup{capbesideposition={left,center},capbesidewidth=4.8cm}}]{figure}[\FBwidth]
    {
         \caption{\textbf{Three-Phased Learning Process} for one of the RNNs trained to emulate the periodic flipflop TA. The learning curves for the network’s time-independent and time-dependent accuracy are plotted in red and blue, respectively, and the learning phases are indicated with the vertical dotted lines. The structure of the learning process was ubiquitous across all networks we trained; the only variations involved the time marking the onset of the phases.}
         \label{fig:PeriodicFFLearningCurve}
    }
    {
        \includegraphics[width=8.5cm]{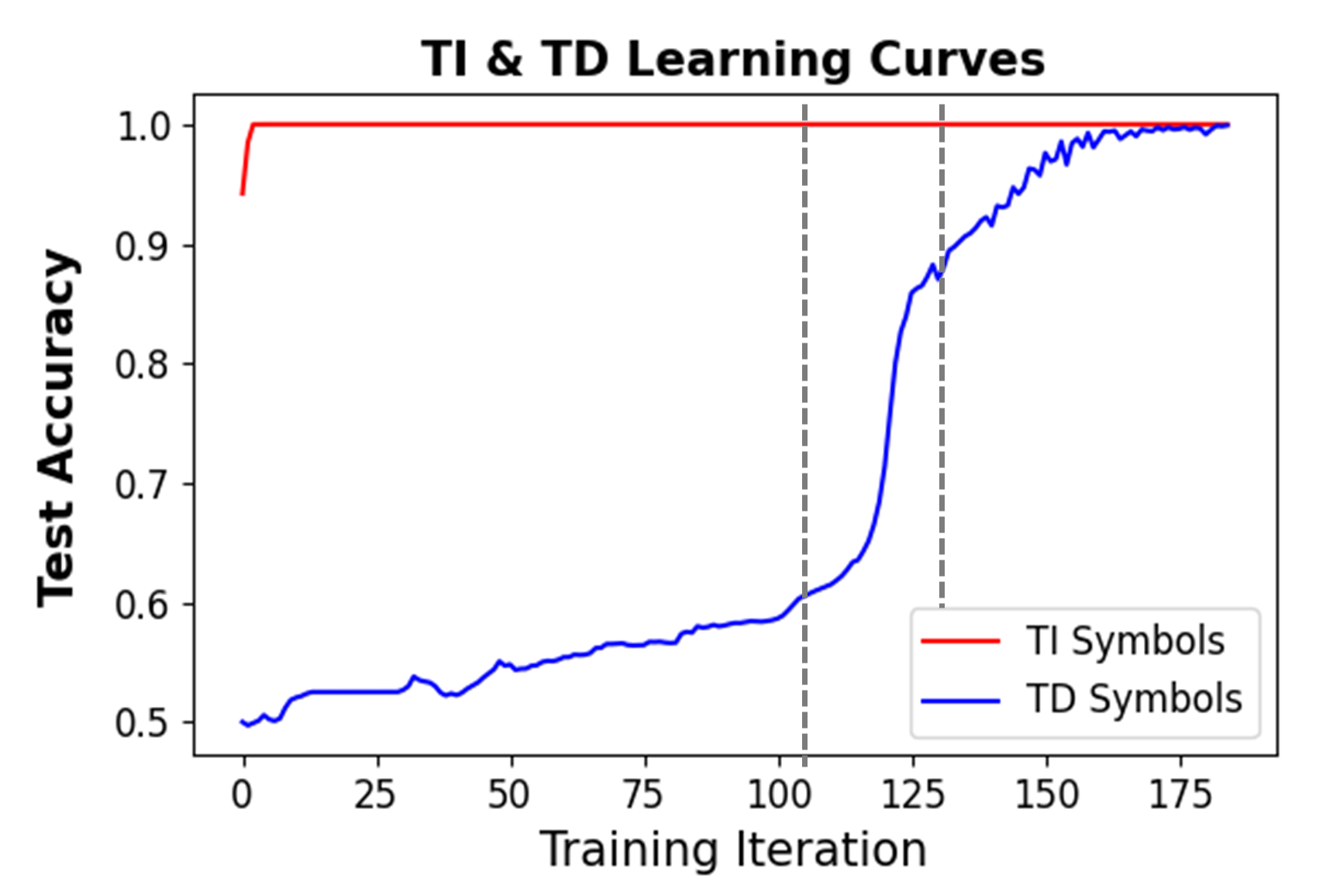}
    }
\end{figure}

For all networks, the time-independent and time-dependent learning curves followed the same three-phased structure, an example of which is shown in Fig. \ref{fig:PeriodicFFLearningCurve}. Training begins with the network perfecting its time-independent behavior, while only slowly improving its time-dependent accuracy above 50\% (equivalent to guessing). This phase lasts for at least half of the training process (on average), after which the network undergoes rapid improvement in its TD accuracy, rising from $<$65\% to $>$90\% in a fraction of the duration of the first phase. The final phase of learning is characterized by a slower convergence of the network’s time-dependent accuracy to $>$99\%.

We emphasize that the structure of the learning process was ubiquitous across all networks trained during our experiment. Experiments using recurrent architectures with gated units exhibit the same learning structure, though the onset of the third phase tends to be earlier for these networks.

\subsection{Periodic Orbits Encoding Time of Day}

We next sought to understand what causes the onset of learning the time-dependent behavior and how this is connected to the time-awareness required by the task. To this aim, we used techniques from dynamical systems theory to explain (1) how the trained network's dynamics encode time of day and (2) how this representation emerges during the training processes.

\begin{figure}[t!]
    \centering
    \begin{subfigure}{0.3\textwidth}
        \includegraphics[width=\textwidth]{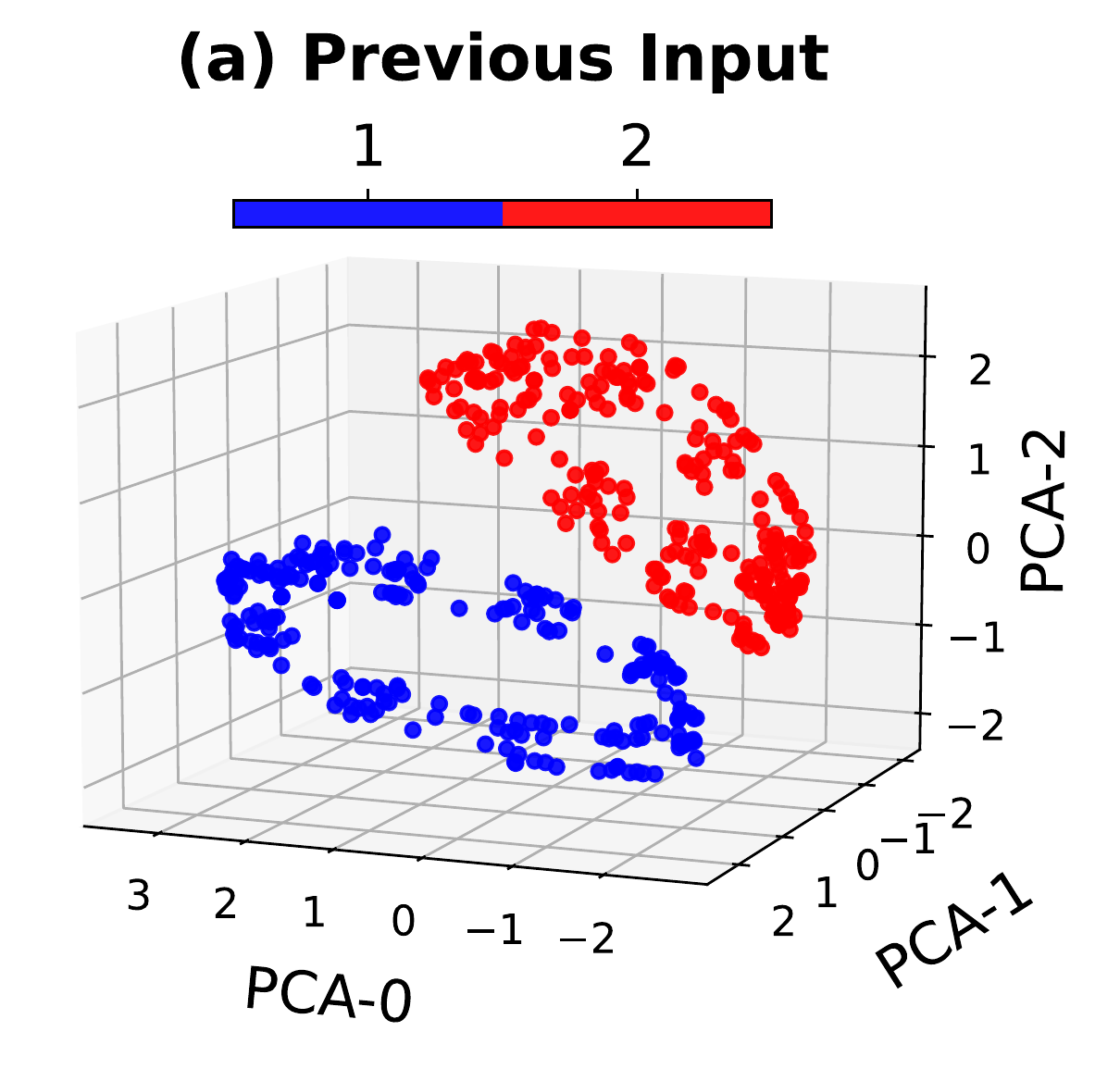}
    \end{subfigure} $\quad$
    \begin{subfigure}{0.3\textwidth}
        \includegraphics[width=\textwidth]{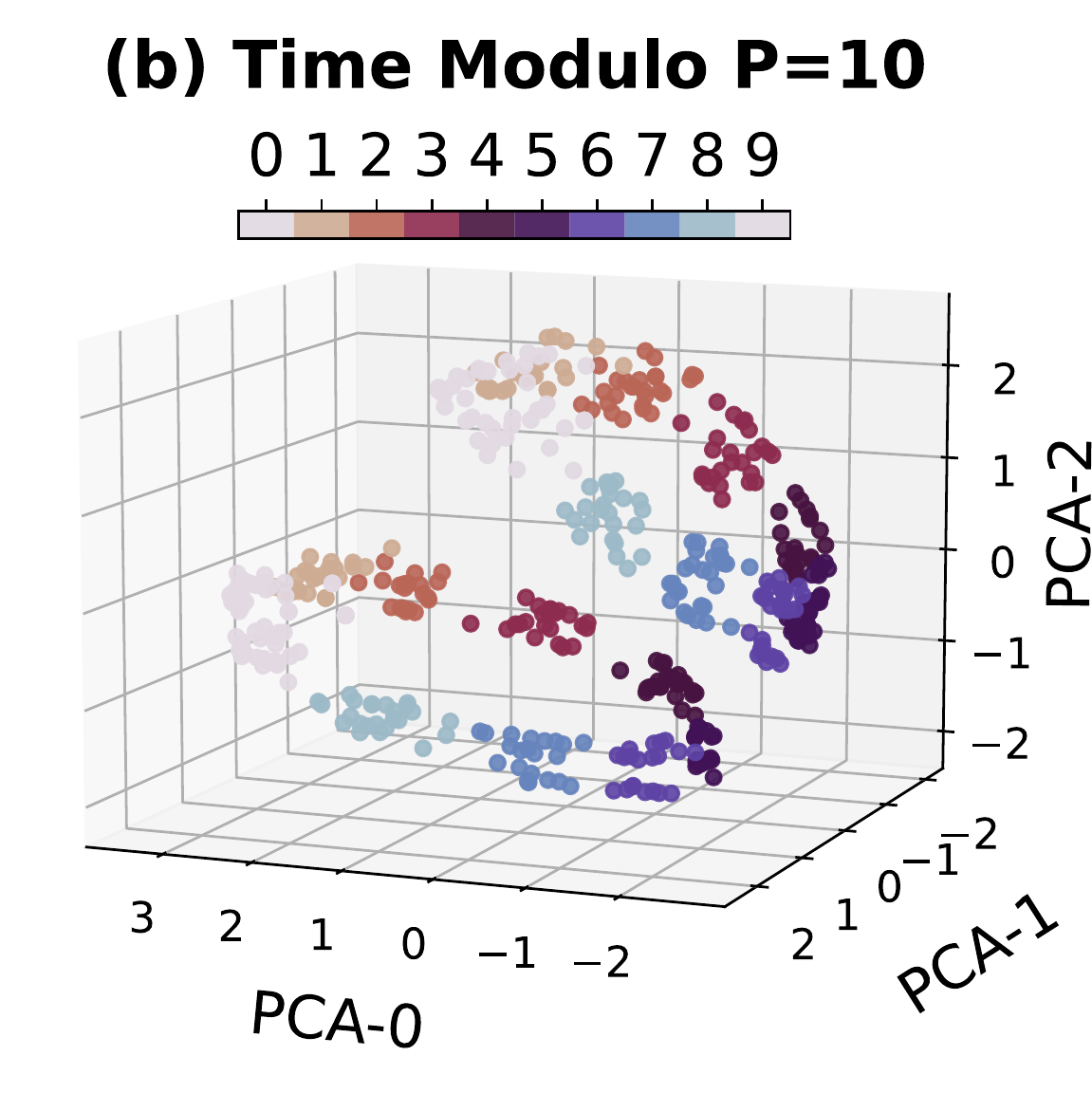}
    \end{subfigure} $\quad$ 
    \begin{subfigure}{0.3\textwidth}
        \includegraphics[width=\textwidth]{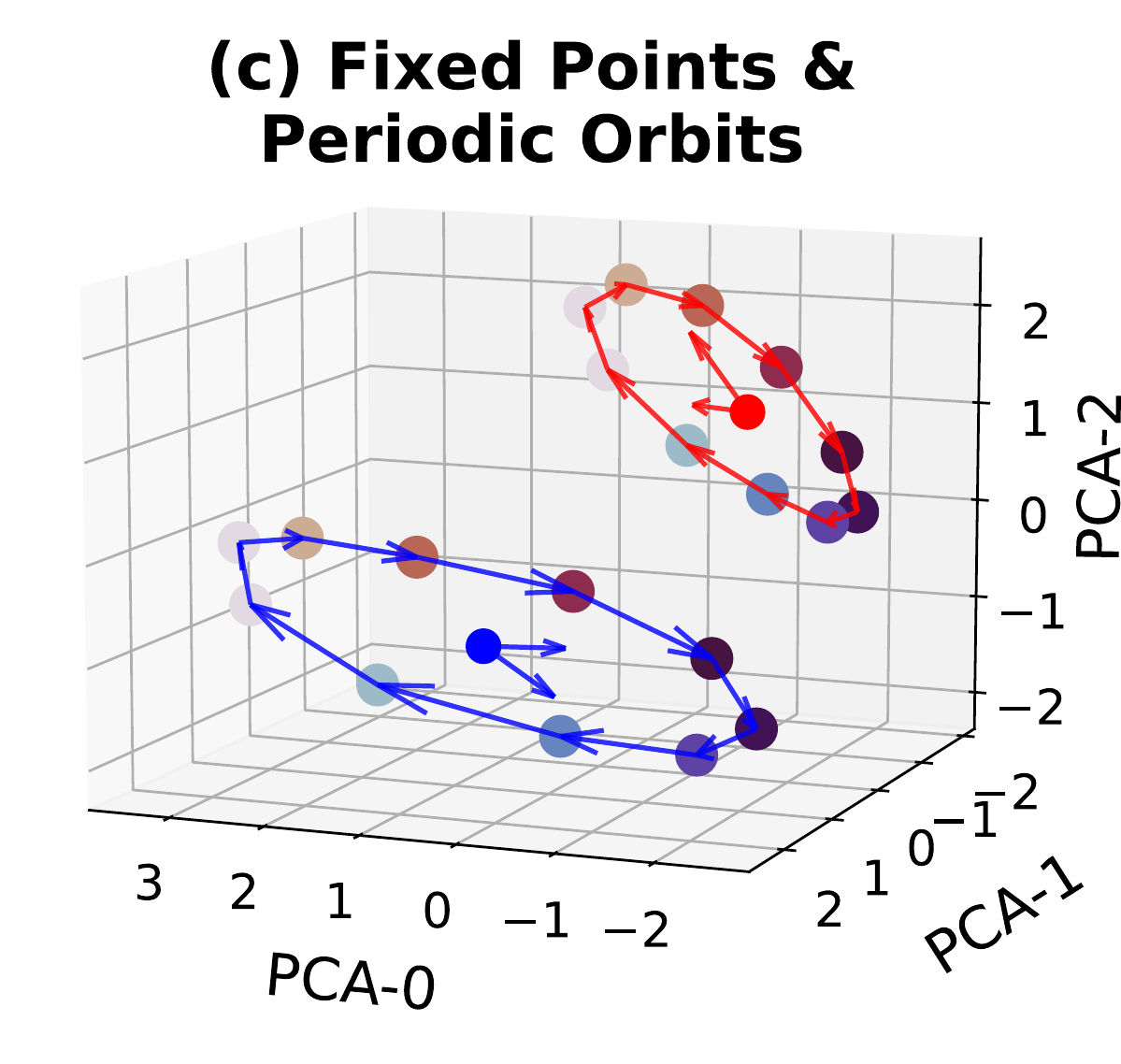}
    \end{subfigure}
    \caption{
        \textbf{PCA Visualization} of the dynamics of an RNN trained to emulate the periodic TF. (a) \& (b) shows the top three principal components of the hidden state dynamics \textit{during inference} on a few example input sequences. They have the same set of points, colored based on different features as indicated. (c) shows the periodic orbits encoding time of day, with each state colored based on time modulo $P = 10$. Refer to Section 4.2 for further details.
    }
    \label{fig:PeriodicFF_PCA}
\end{figure}

Principal component analysis (PCA) of the trained network's dynamics reveals its hidden states during inference organize into two point-clouds resembling rings. These rings encode the two relevant pieces of information required by the network to predict the current TA state: the previous input symbol and the time of day. The rings themselves encode the former data, as indicated in Figure \ref{fig:PeriodicFF_PCA}a, whereas the position along the rings encodes time modulo $P=10$ (Fig. \ref{fig:PeriodicFF_PCA}b). In a sense, the RNN's hidden layer functions as a DFA with $2P$ states, each state corresponding to a pair (previous input, time modulo $P$). The DFA states aren't necessarily specific hidden states; rather, they correspond to general regions of the hidden state space.

The rings themselves cluster about periodic orbits around fixed points (FPs) of the hidden layer. Recall, for a discrete driven dynamical system $h(t+1)=F(h(t),u(t))$, a fixed point of the input $u$ is a state $h_*$ such that $h_*=F(h_*,u)$. Using the FP detection algorithm introduced by \citet{sussillo13}, we found that the networks all had a single unstable fixed point for each input symbol. Figure \ref{fig:PeriodicFF_PCA}c shows the hidden state trajectories induced by constant input strings (e.g. $a...a$ or $b...b$). One can see these are almost perfectly P-periodic, though closer inspection would reveal $h(t)$ and $h(t+P)$ are not perfectly equal. 

\subsection{Bifurcations During Training}

To understand how the input-dependent fixed points emerge through learning, we computed the input-dependent FPs at each training step. We then tracked the stability of these fixed points through training by computing the absolute value $|\lambda_\textrm{max}|$ of the largest eigenvalue of the Jacobian matrix $J_{kl} = \partial F_k(h,u) / \partial h_l$ at the fixed points for each input symbol. Each FP is (locally) stable given constant input if $|\lambda_\textrm{max}| < 1$.

\begin{figure}[b!]
    \centering
    \includegraphics[width=0.8\textwidth]{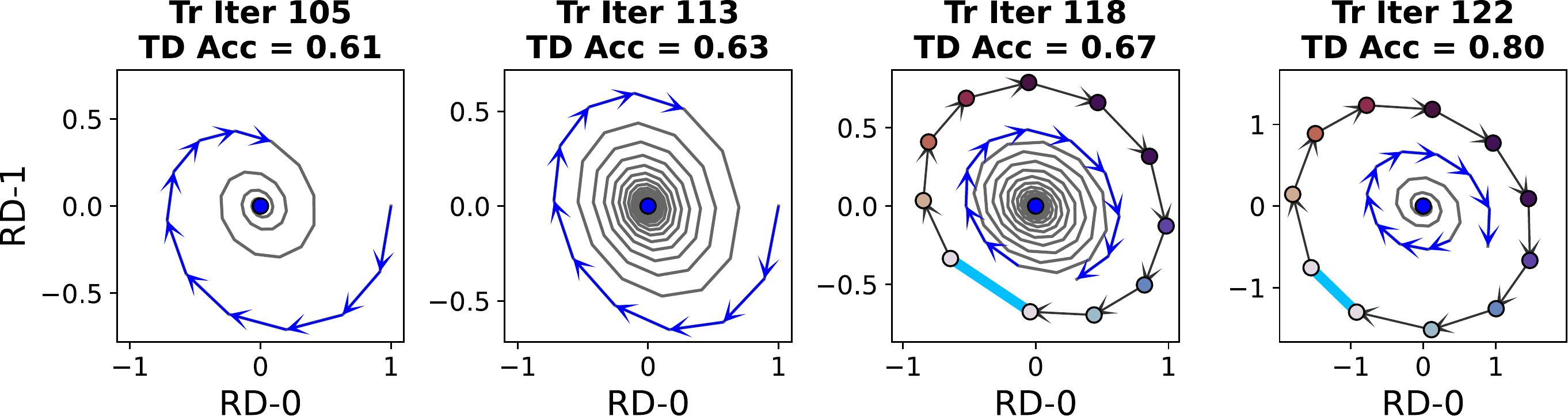}
    \caption{\textbf{Changing dynamics nearby the fixed point for Symbol A during training}. Refer to Section 4.3 for further details.ha A similar bifurcation is observed for the fixed point of Symbol B, but this change in dynamics typically occurs slightly later during training, after the TD accuracy has already escaped from the plateau.}
    \label{fig:PeriodicFFBifurcation}
\end{figure}

We found that all trained networks initially have a single \textit{stable} FP for each input symbol. Early in training, the networks responds to Symbol $b$ (TI) by moving its hidden state closer to the associated FP, near which the networks predicts TA State 2 with a probability close to 1. The networks know that Symbol B is time-independent, and they find a simple strategy to emulate this behavior. The strategy does not indefinitely keep track of time of day, however, because trajectories spiral inwards towards the fixed point. 

Symbol $b$ (TD) also takes the network close to a stable FP during the early stages. Here, we find that the networks' predictions are close to 0.5 probability for both TA states, i.e. it see it sees Symbol $b$ as causing random transitions. In other words, the network has not yet uncovered the hidden periodic variable, so it simply learns to predict the transition probabilities.

The two fixed points remain stable for much of the training process, and this stability results in a period of slow learning in terms of the TD accuracy. As training progresses, we see the emergence of decaying oscillatory dynamics in the hidden state dynamics given fixed input (see Fig \ref{fig:PeriodicFFBifurcation}). The highest magnitude eigenvalue $\lambda_\textrm{max}$ of the Jacobians at the FPs are complex at this point. The subfigures show the projections of the hidden state onto the real and imaginary parts of the associated eigenvectors.

The decay speed decreases as training progresses, which coincides with increasing $|\lambda_\textrm{max}|$ at the fixed point. This largest eigenvalue for each fixed point eventually crosses 1 in absolute value, making the associated fixed point unstable, and at this point we see the emergence of sustained oscillations (given constant input) about the fixed points. Note that theory guarantees that the fixed points are unstable for $|\lambda_\textrm{max}| > 1$, but this instability does not imply the existence of periodic orbits. Our empirical investigation of the networks' dynamics as in Fig. \ref{fig:PeriodicFFBifurcation} verifies that these stable orbits do exist. These trajectories are quasi-periodic, but approach period $P$ as training progresses.

\begin{figure}[t!]
    \centering
    \includegraphics[width=0.85\textwidth]{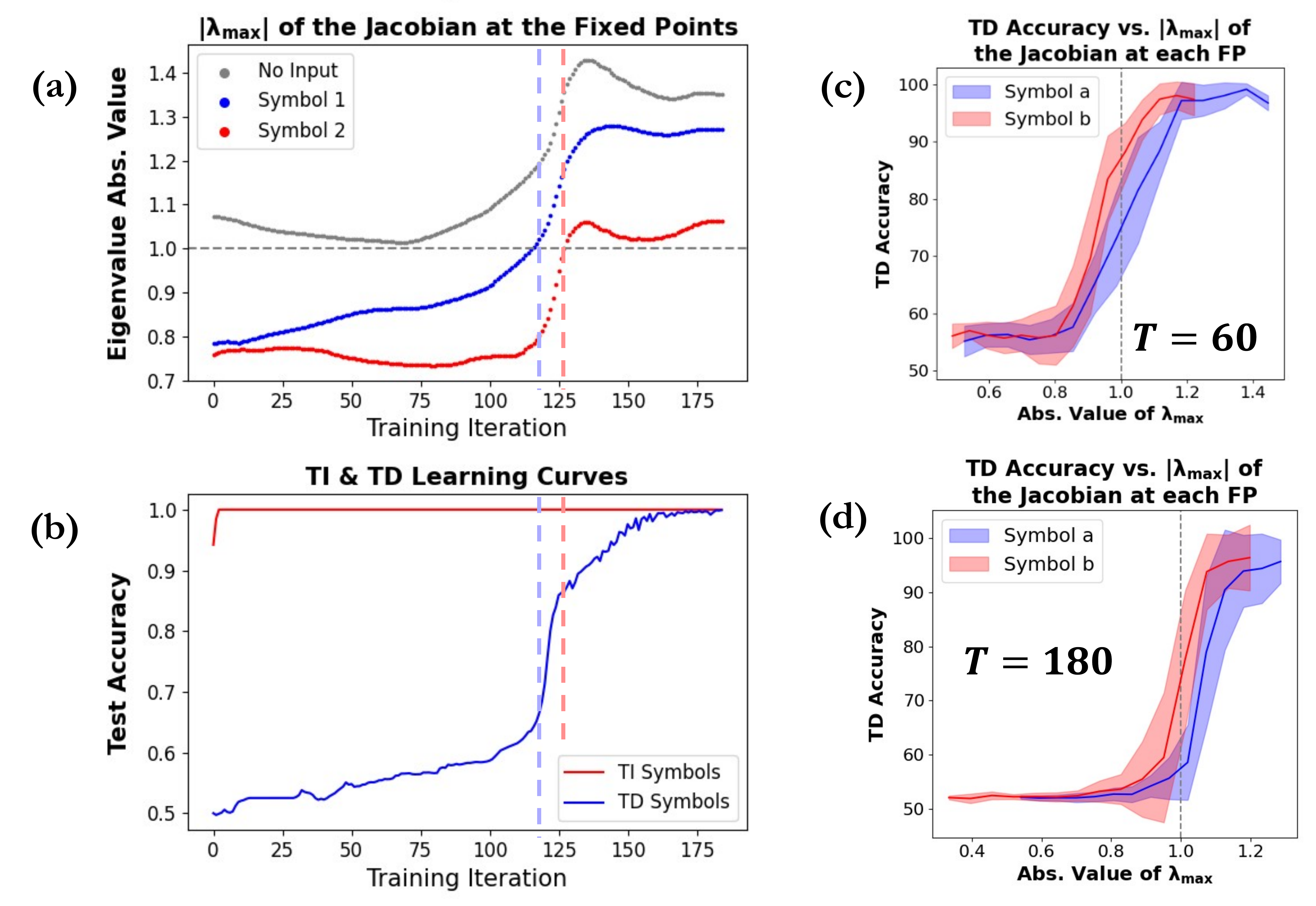}
    \caption{\textbf{Relationship between fixed point stability and TD Accuracy throughout training}. See text immediately below for the details of this figure.}
    \label{fig:PeriodicFF-LCs-and-Lambdas}
\end{figure}

The destabilization of the unstable FP appears to be correlated with the phase of rapid learning of the TD accuracy. In Figure \ref{fig:PeriodicFF-LCs-and-Lambdas}a-b, one can see the plateau in the TD accuracy ends around when the bifurcation point for the TD symbol. Figure \ref{fig:PeriodicFF-LCs-and-Lambdas}c plots the average TD accuracy ($\pm$ 1 stdev) as a function of $|\lambda_\textrm{max}|$ for both Symbols. Across our 15 trials, the average value at which the network's accuracy crosses 75\% is close to $|\lambda_\textrm{max} | = 1$, i.e. precisely when stable periodic behavior emerges about the FPs. This result provides \textit{quantitative} evidence that the cycles are indeed the mechanism learned by the RNNs to encode time.

We found that increasing the training sequence length to $T = 180$ (c.f. Fig. \ref{fig:PeriodicFF-LCs-and-Lambdas}d) increases the average value of $|\lambda_{max}|$ when the TD accuracy crosses 75\%. This result is not unexpected: oscillation period needs to be more precise for the network's predictions to remain accurate over a longer time periods. Still, on average, the TD accuracy does not start to increase until after the FP becomes unstable.

A similar bifurcation is observed by \citet{ribeiro20} in their work on the vanishing and exploding gradient problem, though they do not report a connection between the loss and the bifurcation.

\section{Learning Relative-Timing Temporal Flipflop}

We now extend our analysis to the 2-State Relative-Timing TA constructed using threshold $\tau = 5$ and a design choice probability of seeing a non-null symbol $p = 0.2$. 

\subsection{State Fixed Points Encoding TA States}

\begin{wrapfigure}[34]{r}{0.4\textwidth} 
  \centering
  \includegraphics[width=\textwidth]{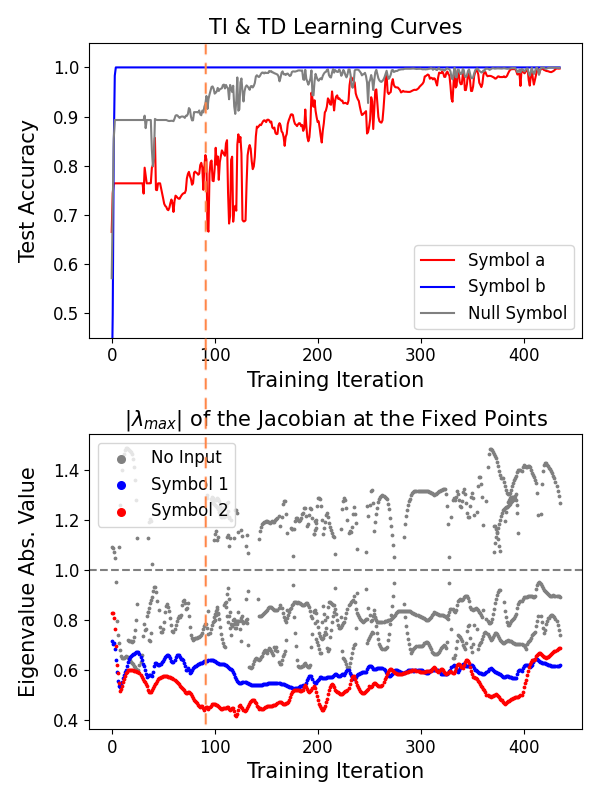}
  \caption{\textbf{The Relative-Timing 3-Phased Learning Process}. Learning dynamics are marked by one stable fixed point ($|\lambda_{max}| < 1$) for each symbol, and one stable null-symbol fixed point for each state---giving rise to $|\Sigma| + |S|$ stable fixed points. The learning starts with just one stable fixed point for the null-symbol, and the second one emerges around iteration 90 as marked by the orange vertical dashed line. The emergence of the second fixed point also marks the exit of the plateau and the beginning of phase three of learning.}    
  \label{fig:relative_timing_lcs}
\end{wrapfigure}

We once again observe three learning phases for the Relative-Timing TA as seen in Figure \ref{fig:relative_timing_lcs}. The time-independent behavior of Symbol \emph{b} is learned almost instantly (10 batches) in comparison to the learning of time-dependent behavior of Symbol \emph{a}. The accuracy associated with the null-symbol input is the mean of the time-independent and time-dependent input symbol accuracy

Figure \ref{fig:relative_timing_lcs} also plots the significant difference in comparison with the Flipflop TA---the emergence of stable ($|\lambda_{max}| < 1$) fixed points for the null symbol $\phi$ and non-null symbols $\{a, b\}$. We notice that there are in total four stable fixed points for the 2-State Relative-Timing TA---one for each symbol $\{a, b\}$ and two for the null-symbol $\phi$. Learning begins with just one stable fixed point for $\phi$, but as it progresses a second one appears as marked by the orange vertical dashed line. The emergence of the second stable fixed point for $\phi$ indicates that the network has started learning its internal representation of time for this task since it coincides with the escape from the learning plateau and the beginning of the third phase of learning. The third phase is incremental in contrast to the rapid learning observed for the periodic TF because the network steadily learns to count up to the exact value of the threshold. When scaled up to an $n$-State TA we can expect $n$ stable fixed points for the null input---one for each state.

This learning process is ubiquitous across all networks trained during our experiment with the variations being in the number of iterations for the second stable null-symbol fixed point to emerge (89 $\pm$ 18, 15 seeds) and the total number of training iterations (452 $\pm$ 103, 15 seeds).

\subsection{Emergence of the Fixed Points During Training}

The emergence of stable fixed points would indicate the lack of oscillations in the dynamics as previously observed for the Flipflop. This is bolstered by the definition of the relative-timing TA in which the time-dependence is not periodic but based on an arbitrary probability value. We hypothesize that we obtain one stable fixed-point with the null-symbol for each state of the TA, and the non-null symbols cause the system to move \emph{away} from these fixed points; the system uses the distance from these stable fixed points to encode the amount of time since a non-null symbol was received, and with each null-input received the system would get closer to these fixed points.

We visualize the RNN hidden state(s) in a lower dimensional subspace to validate this hypothesis. We use a 2-dimensional analysis in which the y-axis indicates some notion of phase/state of the TA and the x-axis encodes a representation of distance (or time) between points projected onto it. We select the y-axis to be the first principal component of the input weight matrix $W_{ih}$ of the last cell in the RNN network, and the x-axis to be the coefficients of a logistic regression model trained to classify whether a given hidden state is above or below threshold (two classes obtained by using the Relative-Timing TA train dataset as an oracle).

\begin{figure}[t!]
    \centering
    \includegraphics[width=\textwidth]{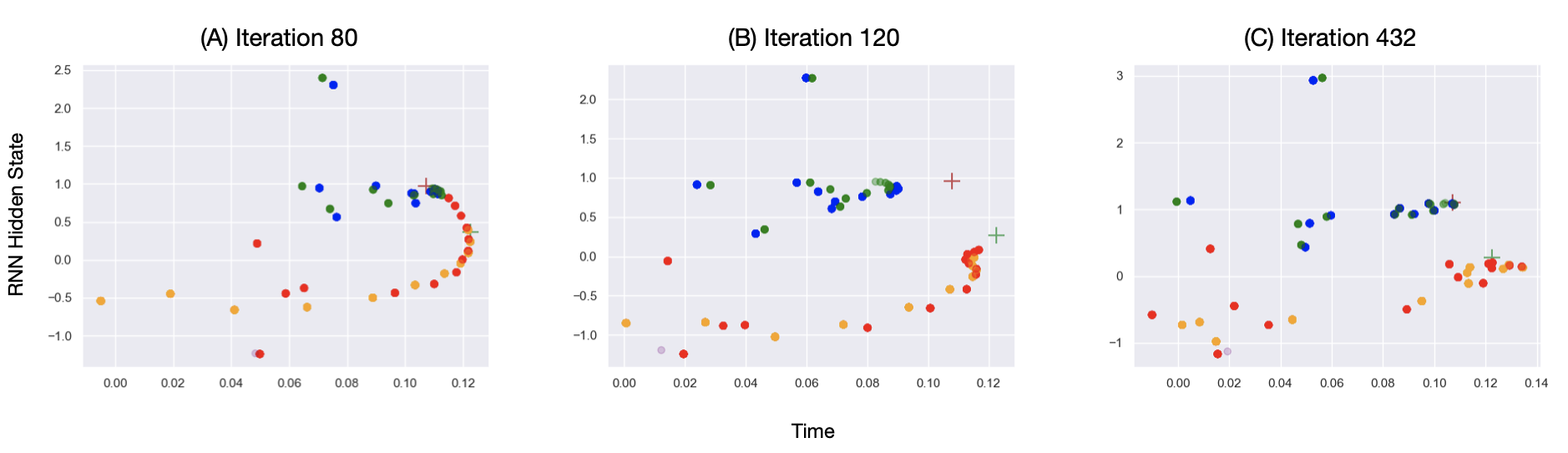}
    \caption{\textbf{Analysis of stable fixed points during training.} There is only one stable FP for the null-symbol in Figure A and all RNN hidden states collapse to it. Figure B is after the emergence of the second null-symbol stable FP marked by the vertical orange line in Figure \ref{fig:relative_timing_lcs}. The dynamics/transitions are now learned and the network incrementally learns to count the latent variable threshold. In Figure C the latent variable threshold is learned, and it takes $\tau$ steps to collapse to the respective stable FP.}
    \label{fig:relative_timing_dynamics}
\end{figure}

Fig. \ref{fig:relative_timing_dynamics} plots this low-dimensional analysis of the RNN hidden state. The two stable fixed points associated with the null-symbol are marked with a cross sign on the right of the plot. The RNN is provided with the input sequence consisting of alternating Symbols \emph{a} and \emph{b} with $2\tau$ null-symbols between them, e.g. $a, \phi, ..., \phi, b, \phi, ..., \phi, a$ This allows us to study the convergence to stable fixed points and how the TA state behaves with each symbol as an input. We only change the color of the RNN state when a non-null symbol is received.

We observe it takes the threshold $\tau$ timesteps for the RNN hidden state to collapse to one of the stable fixed points. The stable fixed point it collapses to depends on the current state of the TA network---the red cross corresponds to State 1 and the green cross to State 2. When provided with a non-null symbol, the RNN state is pulled away from the null fixed point and towards the left of the plot. Depending on which symbol was received, the RNN state is pulled to a specific point: the upper region ($> 0$) containing the green and blue states are when Symbol \emph{a} is received and the lower region ($< 0$) containing the red and orange states are for Symbol \emph{b}.

We also note that during learning when there is only 1 null-symbol fixed point (Figure \ref{fig:relative_timing_dynamics}-A), then irrespective of the current state there is always a collapse to it. This is because the network hasn't learned to count to the threshold and is missing any understanding of the time-dependence. Figure \ref{fig:relative_timing_dynamics}-B represents the emergence of the second fixed point (pass the vertical orange line in Figure \ref{fig:relative_timing_lcs}) and hence there is an increase in the accuracy. However, while the dynamics are learned, the representation of time and the ability to count still needs to be learned. This process of fine-tuning can take longer, thereby resulting in a steady learning curve until the network learns to count and converge to a respective stable fixed point in $\tau$ steps as seen in Figure \ref{fig:relative_timing_dynamics}-C.

\section{Discussion}

In this paper, we provide one of the first detailed studies of recurrent networks trained to infer hidden temporal variables. Using tools from dynamical systems, we explain the ways that these networks represent time in their hidden states and present evidence to illuminate how these internal representations emerge \textit{during} training. We accomplish this by defining a new family of automata-based time-aware sequence modeling tasks. For the state-independent automata we call Temporal Flipflops (TF), we find that the networks learn reusable behaviors of time that improve learning and generalization. Our experiments show that a common three-phase learning structure is observed independent of the form of time-dependence (periodic vs. time counting), but we observed differences in the emergent temporal representations and their associated training bifurcations.

Learning curve plateaus can be seen a wide range of ML applications \citep{Behera2021, Bemporad2021, Jiang2019, Kim2020, Murakami2022, Tang2021, Vecoven2021}, so this style of analysis may provide insights into the training pathologies of recurrent architectures in other contexts. This plateauing phenomena is closely related to vanishing and exploding gradients, a fundamental challenge in learning long-term dependencies with RNNs. There is a long-standing hypothesis attributing gradient pathologies during training to bifurcations in RNN dynamics \citep{Doya1993, bengio94, Pascanu2013}, and to our knowledge, this paper is one of the first to demonstrate this connection empirically for trained RNNs with 100s of parameters. Future studies using dynamical systems to analyze the training process itself may prove to be informative to designing architectures that more efficiently learn long-term dependencies.

Despite the promise of dynamical systems to elucidate the learning process, there is no guarantee that this analysis will be applicable to recurrent models trained on other tasks, even those described by timed automata. Preliminary experiments suggest that the long-term dependencies required to learn state-dependent TA extend the learning process even further and may make it computationally infeasible to perform fixed point stability analysis with high temporal granularity. For this form of analysis to be applied more generally, there is a need for more efficient methods of tracking the fixed points of the RNNs through training, for instance by modifying the learning process to make the fixed points easier to track \citep{Smith2021}. Still, even without these developments, exciting directions for future work include temporal automata with forms of time-dependency, larger state spaces, state-dependent transitions, and multiple types of time-dependence.

\bibliographystyle{unsrtnat}
\bibliography{TA_RNNs}

\section*{Supplemental Material}
\appendix
\setcounter{figure}{0}
\renewcommand\thefigure{S\arabic{figure}}    

\section{Experiment Details}

\subsection{Architecture}
We used single layer Vanilla RNNs for both flipflop automata tasks. Given an input sequence $(u(1), u(2), ..., u(T))$ with each $u(t)\in \mathbb{R}^{N_{in}}$, an Elman RNN produces the output sequence $y = (y(1),...,y(T))$ with $y(t) \in \mathbb{R}^{N_{out}}$ following the equations
\begin{equation}
h(t+1) = \tanh(W_{hh} h(t) + W_{uh} u(t) + b_h) \quad,\quad y(t) = \sigma(W_{hy} h(t) + b_y)
\end{equation}
Here $W_{uh} \in \mathbb{R}^{N_h \times N_{in}}$, $W_{hh} \in \mathbb{R}^{N_h \times N_h}$, and $W_{hy} \in \mathbb{R}^{N_{out} \times N_h}$ are the input, hidden, and output weights, respectively. These parameters were initialized with each component randomly drawn from a normal distribution with mean 0 and standard deviation $1/\sqrt{N_{in}}, \ 1/sqrt{N_h}, 1/\sqrt{N_{out}}$ respectively.

$N_h$ is the dimension of the hidden state space, which we set to $N_h = 32$ for both the periodic and relative timing temporal flipflop experiments.

The parameters $b_h \in \mathbb{R}^{N_h}$ and $b_y \in \mathbb{R}^{N_{out}}$ are bias vectors.  These vectors were also initialized with each component randomly drawn from a normal distribution with mean 0 and standard deviation $1/\sqrt{N_{h}}$ and $1/\sqrt{N_{out}}$ respectively.

The initial hidden state $h(0) \in \mathbb{R}^{N_h}$ for each model was also trained parameter. The components of these vectors were initially drawn from a normal distribution with mean 0 and standard deviation $g_{h_0} / \sqrt{N_h} $ with $g_{h_0} = 0.05$.

\subsection{Task Parameters}
For both tasks, we generated a dataset with 32768 training examples and 4096 test examples. Each example was a pair of sequences $(u, y)$ with sequence length $T = 60$ where $u = (u(1),u(2),...,u(T))$ is a sequence of TA input symbols and $y = (y(1),y(2),...,y(T))$ is the associated TA output sequence. The flipflop TA studied in this paper both had two input symbols, which where represented as vectors using 1-hot encoding: $u(t) = [1,0]^\top$ to indicate Symbol A and $u(t) = [0,1]^\top$ to indicate Symbol B. These TA also had two states, which we represented using a single binary value: $y(t) = 0$ and $y(t) = 1$ States 1 and 2 of the TA, respectively.

\begin{table}[t!]
\begin{center}
\begin{tabular}{| c | c | c |}
\hline
 & Periodic Timing & Relative Timing \\ \hhline{|=|=|=|}
 Input \& output dimensions $N_{in}, N_{out}$ & 2,1 & 2,1 \\ \hline
 Hidden dimension $N_h$ & 32 & 32 \\ \hline
 Input probability $p$ & 1.0 & 0.2 \\ \hline
 TA period $P$ & 10 & --- \\ \hline
 TA timing threshold $\tau$ & --- & 10 \\ \hline
 Training sequence length $T$ & 60 & 60 \\ \hline
 Max training iterations & 512 & 512 \\ \hline
 Learning rate & $10^{-3}$ & $5\cdot 10^{-4}$ \\ \hline
 (Mini)batch size & 128 & 128 \\ \hline
 RNN Optimizer & Adam & RMSProp \\ \hline
\end{tabular}
\caption{
Architecture, Task, \& Training Parameters\\
NOTE: A training \textit{iteration} is different from an \textit{epoch}, which consists of as many training iterations required to loop through the dataset once. With a batch size of 128 and a training dataset with $2^{15}$ examples, each epoch consists of $256$ training iterations. Our models required only 1-2 epochs to converge.
}
\label{table:training-params}
\end{center}
\end{table}

\subsection{Training Parameters}
We used PyTorch's \citep{Paszke2019} implementation of Vanilla RNNs (the \textsc{RNNCell} class in particular). The models were trained using Adam optimization \citep{Kingma2014} for the periodic flipflop TA and RMSProp \citep{Graves2013} for the relative timing TA. Both tasks used binary cross entropy loss, but similar results were obtained for MSE loss as well. 

We manually tuned training hyperparameters for each task. The values used for our experiments are listed in Table \ref{table:training-params}. 

\subsection{Fixed Point Computation}

We used our own implementation of fixed point finding algorithm for RNNs introduced by \citet{sussillo13}. An implementation of their algorithm exists for the Tensorflow machine learning library \citep{Golub2018}, but we found it easier to implement our own version of the code than to transition our work from PyTorch to Tensorflow.

\section{Visualization of Relative-Timing TA}

In order to visualise the hidden states of the network trained on the relative-timing TA and validate our hypothesis, we project into a two-dimensional space. We aim to have the y-axis represent the state of the TA and the x-axis encode a representation of distance (or time) between points.

Accordingly, the y-axis is constructed by conducting PCA and selecting the first principal component of the weight matrix $W_{ih}$ multiplied to input vectors of the last RNN cell. We call this vector $h_{TA} \in R^{d}$ since it encodes the RNN representation of the TA state, where $d$ is the size of the RNN hidden state. To obtain the x-axis and represent distance, we train a logistic regressor to classify RNN hidden states into binary outputs corresponding to the latent variable $\Theta_\textrm{relative}$ defined in Section \ref{sec:TF-Time-Dependence}. Training a classifier provides an implicit measure of distance from the boundary between classes, something we notice when we project our points in lower dimensional spaces. We call the weights of the classifier $h_{lr} \in R^d$.

We now want to project the RNN hidden states into $Span\{h_{TA}, h_{LR}\}$. To do so we construct matrix $H = [h_{TA}^T, h_{LR}^T] \in R^{d \times d}$. 

Using linear algebra, the projection of a given hidden state $h$ can be obtained by $(w_{TA}, w_{lr}) = H^+ h$ where $H^+ = (H^TH)^{-1}H^T$ and $(w_{TA}, w_{lr})$ are the desired coordinates in the span of the vectors.

\section{Additional Results: Periodic Flipflop}

In this appendix we include additional results from our experiment involving the periodic flipflop automaton. These results were generated using the same procedure and parameters, with the difference arising solely from the number used to seed the random number generator.

\clearpage
\subsection{Post-Training Dynamics}

\begin{figure}[h!]
    \centering
    \begin{subfigure}{0.3\textwidth}
        \includegraphics[width=\textwidth]{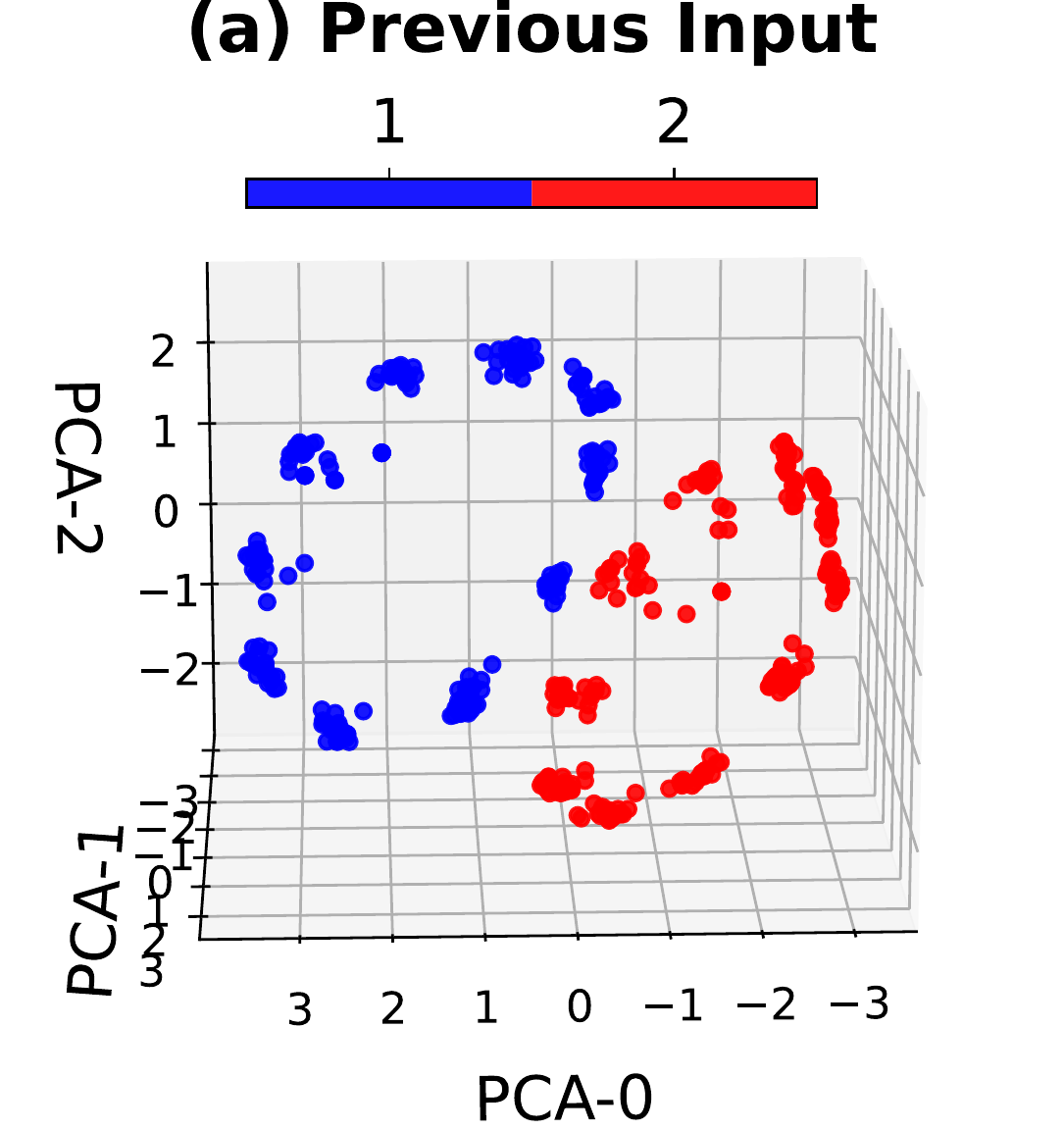}
    \end{subfigure} $\quad$
    \begin{subfigure}{0.3\textwidth}
        \includegraphics[width=\textwidth]{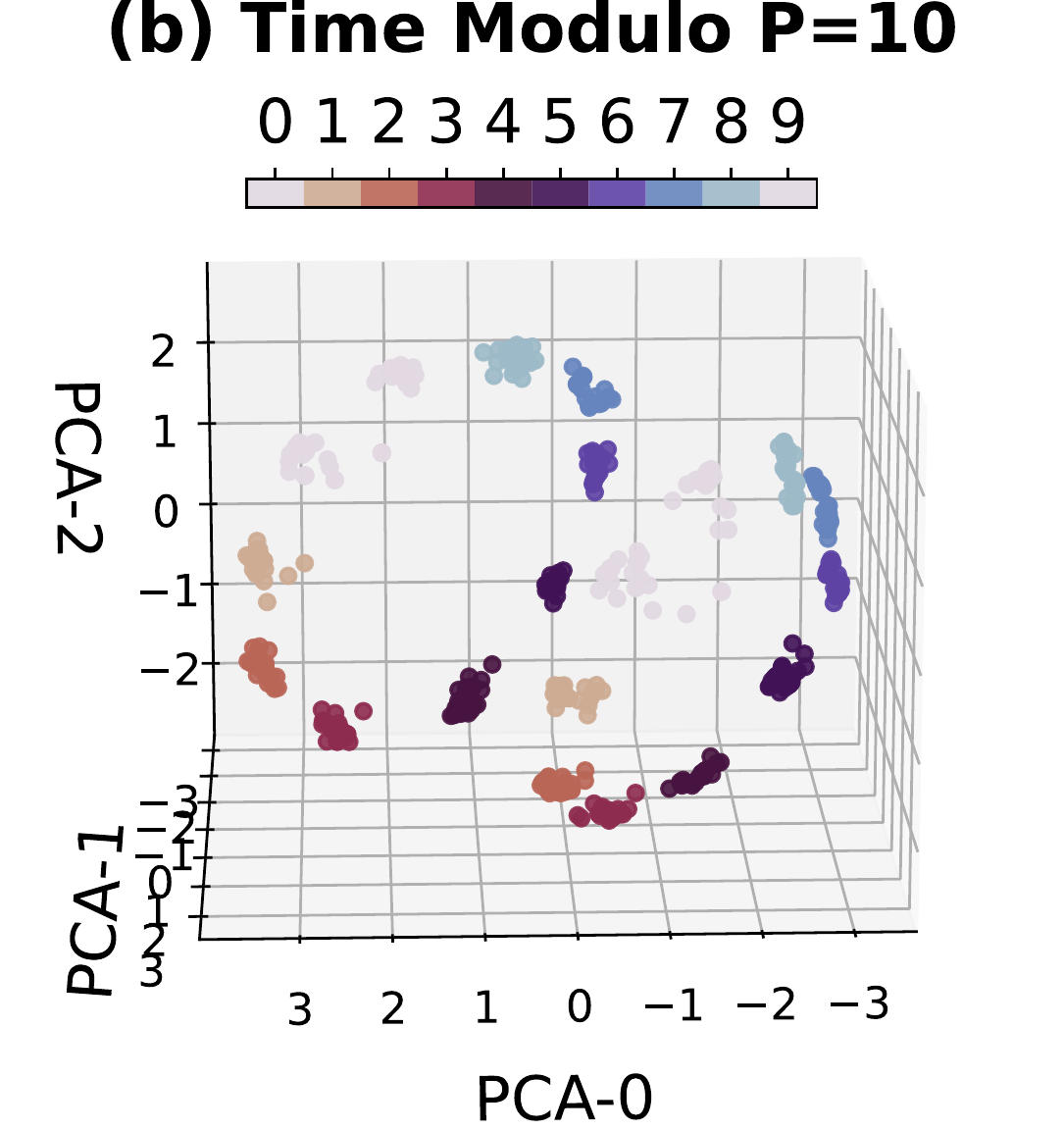}
    \end{subfigure} $\quad$ 
    \begin{subfigure}{0.3\textwidth}
        \includegraphics[width=\textwidth]{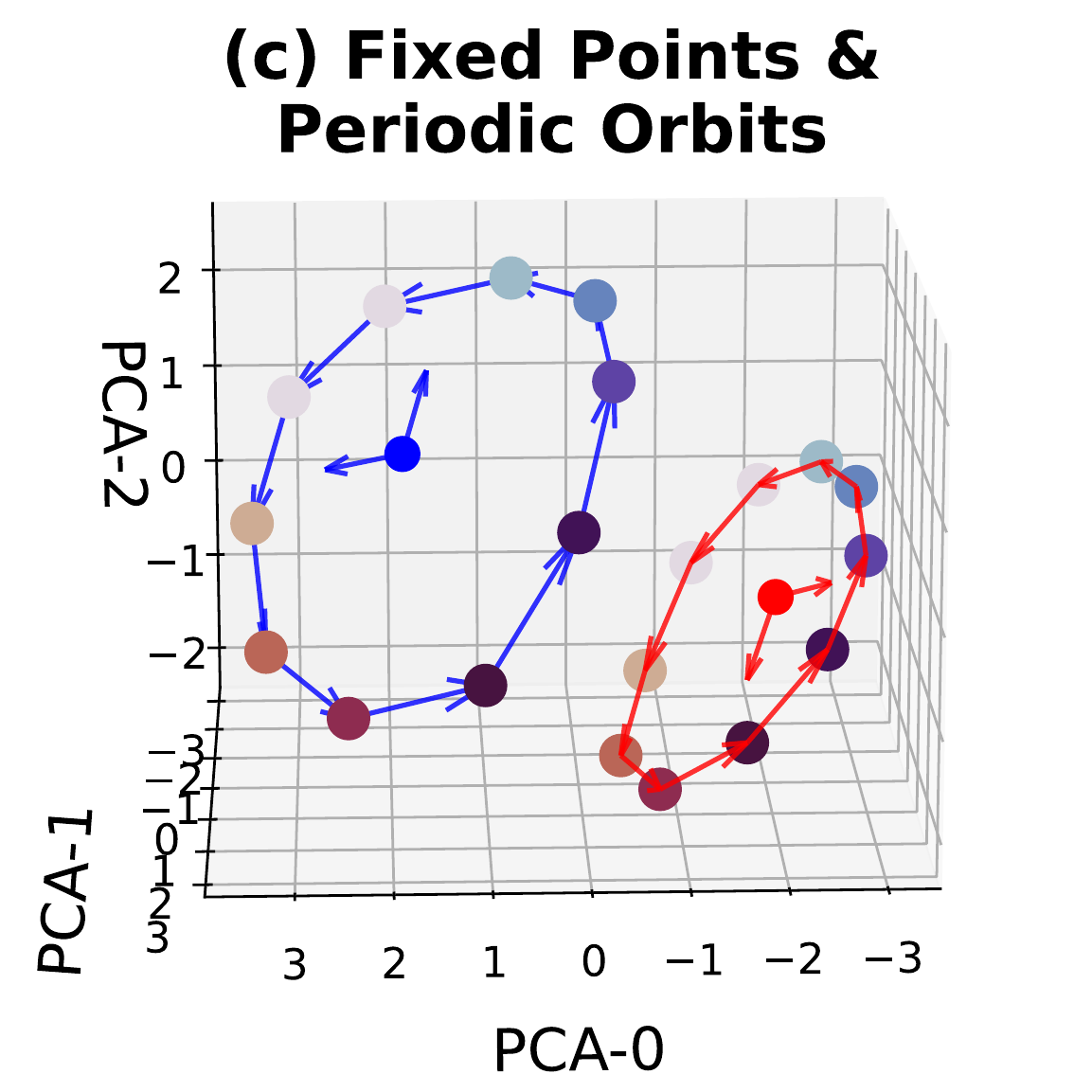}
    \end{subfigure}
    \caption{
        \textbf{PCA Visualization of the dynamics of an RNN trained to emulate the periodic TF} --- a second example, akin to Fig. \ref{fig:PeriodicFF_PCA}. This figure shows the post-training dynamics of a network trained using the same procedure and parameters described in Section B of the Supplemental Material, just with a different random seed.\\
        (a) \& (b) show the top three principal components of the hidden state dynamics \textit{during inference} on a few example input sequences. They have the same set of points, colored based on different features as indicated. (c) shows the periodic orbits encoding time of day, with each state colored based on time modulo $P = 10$.\\
        The PCA of the trained network's dynamics reveals its hidden states during inference organize into two point-clouds resembling rings. These rings encode the two relevant pieces of information required by the network to predict the current TA state: the previous input symbol and the time of day. The rings themselves encode the former data, as indicated in subplot (a), whereas the position along the rings encodes time modulo $P=10$ (subplot b).\\
        The rings themselves cluster about periodic orbits around fixed points (FPs) of the hidden layer. After training, we found that the networks all had a single unstable fixed point for each input symbol, indicated by the blue and red points in subplot (c), which also shows shows the hidden state trajectories induced by constant input strings (e.g. $a...a$ or $b...b$).
    }
    \label{fig:Supp_PeriodicFF_PCA}
\end{figure}

\begin{figure}[h!]
    \centering
    \includegraphics[width=0.8\textwidth]{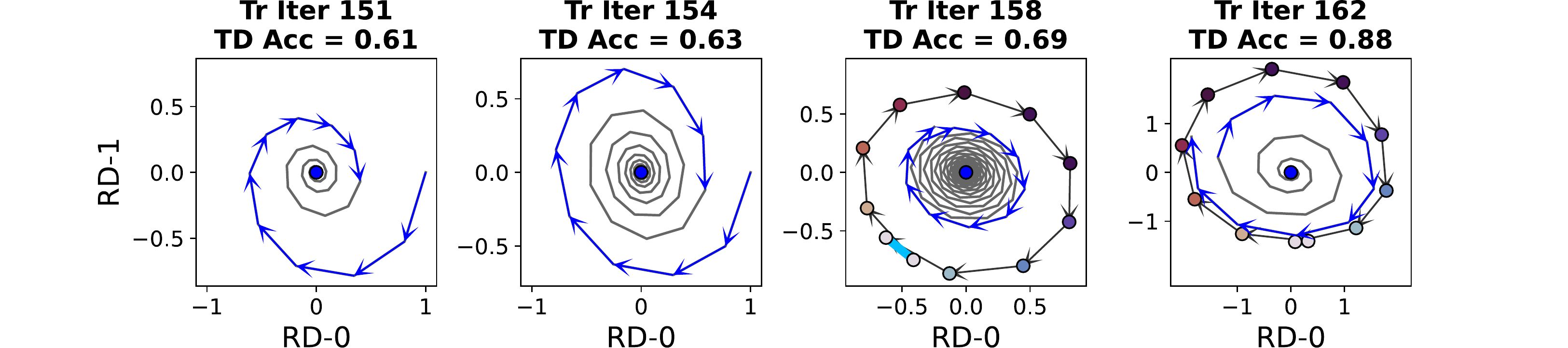}
    \caption{\textbf{Changing dynamics nearby the fixed point for Symbol A during training} --- a second example, akin to Fig. \ref{fig:PeriodicFFBifurcation}.\\ 
    We found that all trained networks initially have a single \textit{stable} FP for each input symbol. As training progresses, we see the emergence of decaying oscillatory dynamics in the hidden state dynamics given fixed input, as shown in the left two plots of this figure. The highest magnitude eigenvalue $\lambda_\textrm{max}$ of the Jacobians at the FPs are complex at this point. The plots show the projections of the hidden state onto the real and imaginary parts of the associated eigenvectors. \\
    The decay speed decreases as training progresses, which coincides with increasing $|\lambda_\textrm{max}|$ at the fixed point. This largest eigenvalue for each fixed point eventually crosses 1 in absolute value, making the associated fixed point unstable, and at this point we see the emergence of sustained oscillations (given constant input) about the fixed points. These trajectories are quasi-periodic, but approach period $P$ as training progresses.}
\end{figure}

\clearpage
\subsection{Time-Dependent Accuracy vs. Fixed Point Stability}

\begin{figure*}[h]
\centering
\begin{subfigure}{0.46\textwidth}
    \centering
    \includegraphics[width=\textwidth]{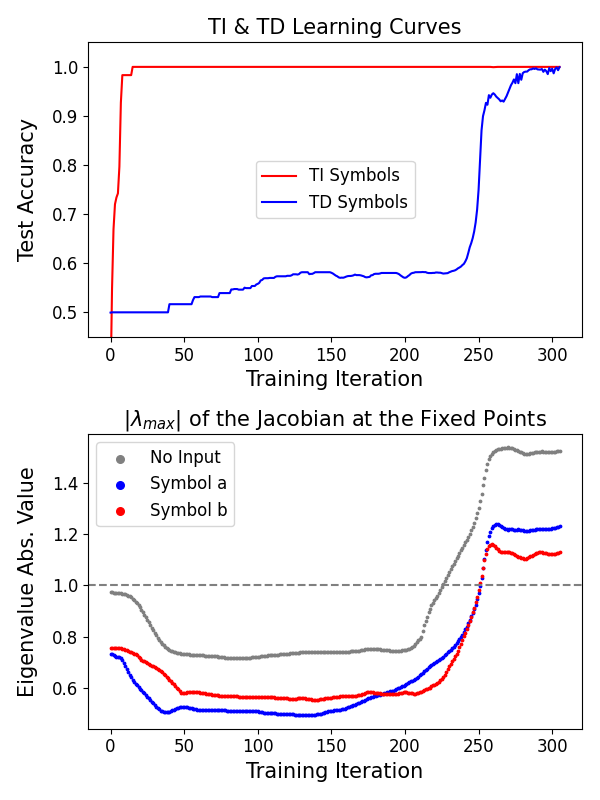}
    \caption{}
\end{subfigure} $\qquad$
\begin{subfigure}{0.46\textwidth}
    \centering
	\includegraphics[width=\textwidth]{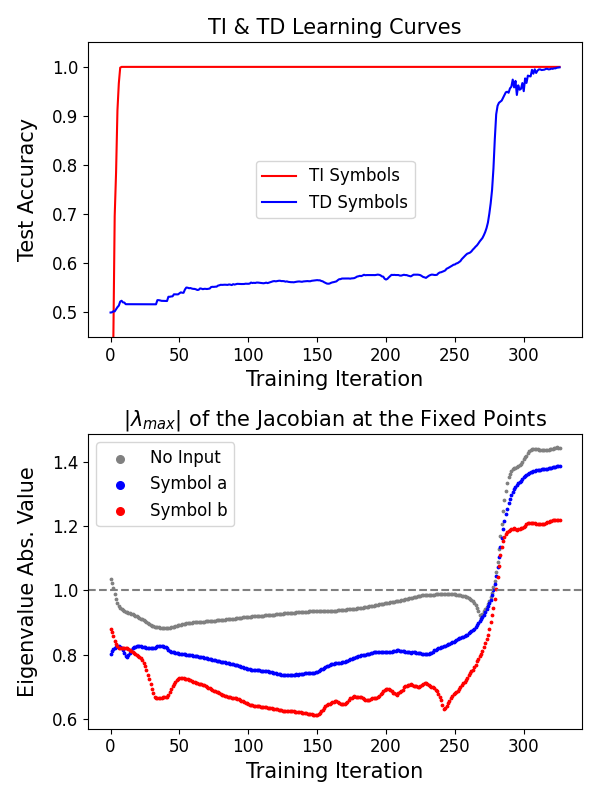}
    \caption{}
\end{subfigure}
\caption{ \textbf{Relationship between fixed point stability and TD Accuracy throughout training} --- further examples for RNNs trained with sequence length $T = 60$. (a) and (b) refer to two different networks trained using the same procedure and parameters described in Section B of the Supplemental Material, just with different random seeds.\\
As in Section \ref{PeriodicFF}, the time-independent and time-dependent learning curves (top row of plots) for these networks followed the same three-phased structure. Training begins with the network perfecting its time-independent behavior, while only slowly improving its time-dependent accuracy above 50\% (equivalent to guessing). This phase lasts for at least half of the training process (on average), after which the network undergoes rapid improvement in its TD accuracy, rising from $<$65\% to $>$90\% in a fraction of the duration of the first phase. The final phase of learning is characterized by a slower convergence of the network’s time-dependent accuracy to $>$99\%.\\
The bottom row of plots tracks of the modulus $|\lambda_\textrm{max}|$ of the largest eigenvalue $\lambda_\textrm{max}$ for all fixed points (FPs) of the networks. Each of these networks had a \textit{single} fixed point associated with each type of input --- Symbol A, Symbol B, and the ``Null Symbol'' (the zero vector). The FPs associated with Symbols A and B are initial \textit{stable} fixed points because their $|\lambda_\textrm{max}| < 1$, but they \textit{destabilized} later in training when their $|\lambda_\textrm{max}|$ crosses the threshold of 1. The destabilization happens at the same time for both Symbol A and Symbol B here, though this result does not hold in general.\\
The destabilization of the FPs of Symbols A and B appears to be correlated with the phase of rapid learning of the TD accuracy. One can see the plateau in the TD accuracy ends around when the bifurcation point for the TD symbol. \\
The Null Symbol also has a single fixed point for both of these networks. This fixed point is stable early in training and becomes unstable as training progresses, but the point of destabilization does not always coincide with spikes in the network's TD accuracy.
}
\end{figure*}

\clearpage
\begin{figure*}[t!]
\centering
\begin{subfigure}{0.46\textwidth}
    \centering
    \includegraphics[width=\textwidth]{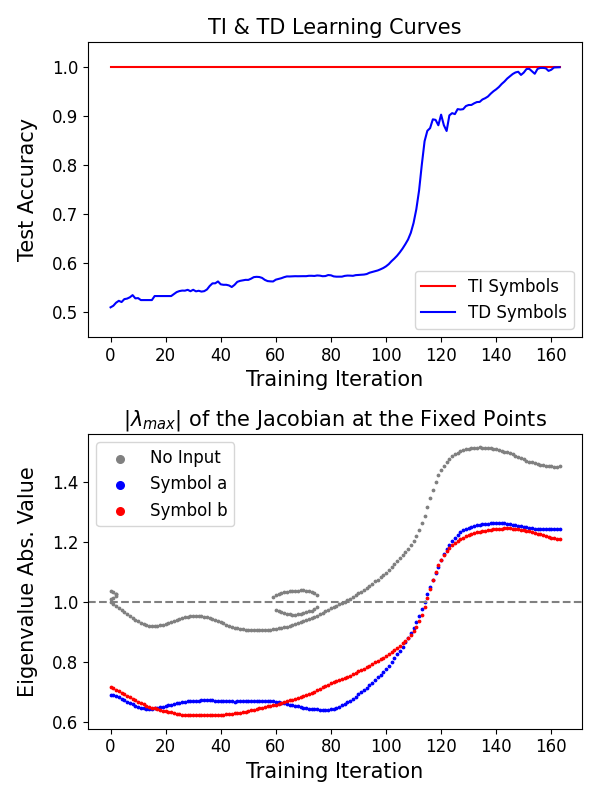}
    \caption{}
\end{subfigure} $\qquad$
\begin{subfigure}{0.46\textwidth}
    \centering
	\includegraphics[width=\textwidth]{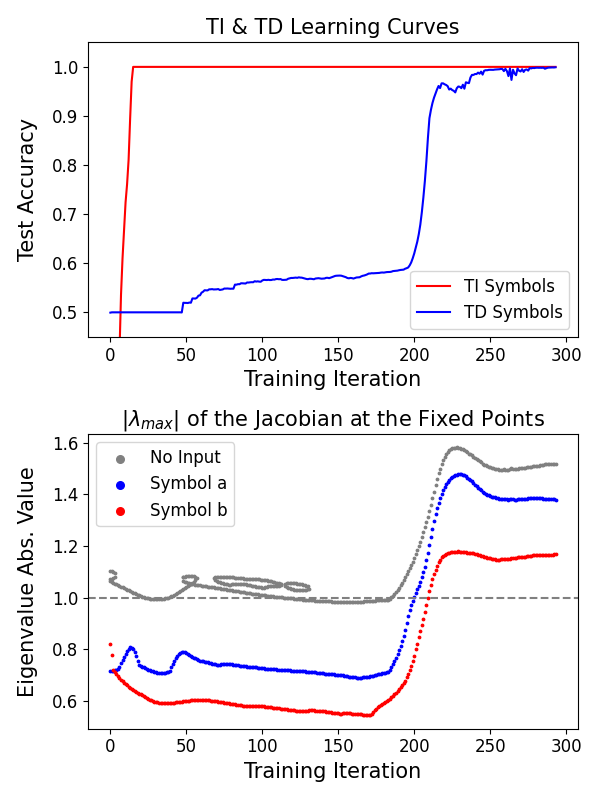}
    \caption{}
\end{subfigure}
\caption{ \textbf{Relationship between fixed point stability and TD Accuracy throughout training} --- further examples for RNNs trained with sequence length $T = 60$. (a) and (b) refer to two different networks trained using the same procedure and parameters described in Section B of the Supplemental Material, just with different random seeds.\\
The learning curves for these plots are qualitatively the same as those in the figure on the previous page. Similarly, the correlation between the increase in the networks' time-dependent accuracy (top row, in blue) and the moduli $|\lambda_\textrm{max}|$ of the largest eigenvalue of the fixed point (FP) associated with Symbol A (bottom row, in blue). \\
Here, we do see a difference in the structure of the fixed points associated with Null Symbol (bottom row, in gray): For both of these networks, a pair of fixed points associated with the null symbol appears during training and vanishes before the networks' converge. This result may not be illuminating for the periodic flipflop task, in which the network was never presented with the null symbol as input.
}
\end{figure*}

\clearpage
\begin{figure*}[t!]
\centering
\begin{subfigure}{0.46\textwidth}
    \centering
    \includegraphics[width=\textwidth]{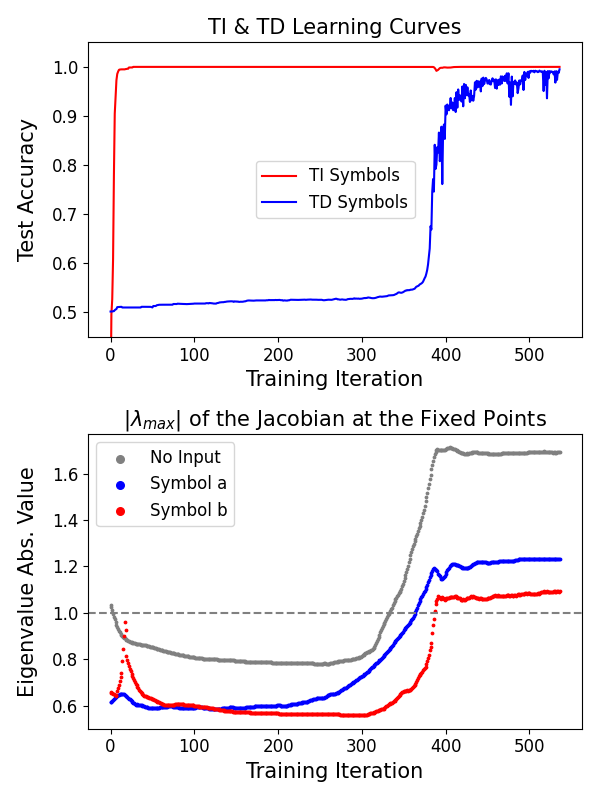}
    \caption{}
\end{subfigure} $\qquad$
\begin{subfigure}{0.46\textwidth}
    \centering
	\includegraphics[width=\textwidth]{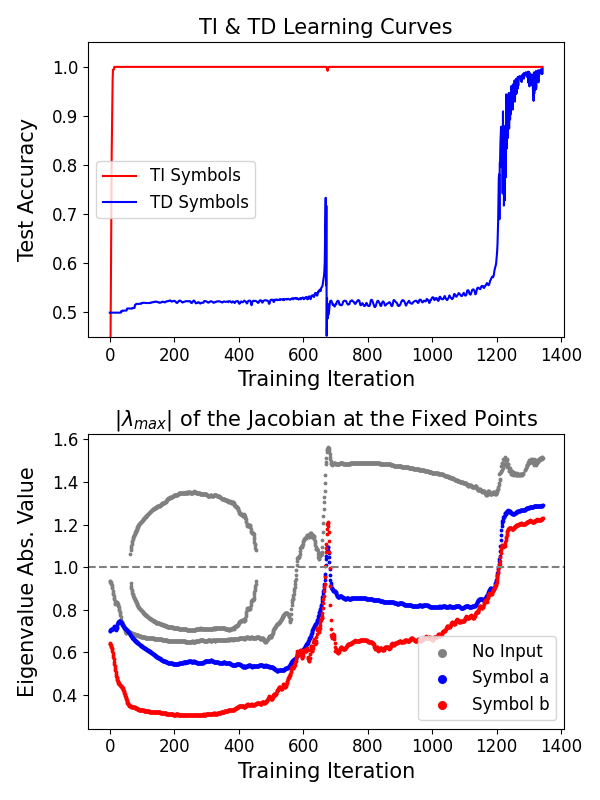}
    \caption{}
\end{subfigure}
\caption{ \textbf{Relationship between fixed point stability and TD Accuracy throughout training} --- examples for RNNs trained with sequence length $\mathbf{T = 180}$. (a) and (b) refer to two different networks trained using the same procedure and parameters described in Section B of the Supplemental Material, just with $T = 180$, more training iterations, and different random seeds.\\
The learning curves for these plots are qualitatively the same as those in the figures on the previous two page. Quantitatively, these networks trained on longer sequence lengths require more training iterations to converge, which is reasonable because these models were evaluated over longer periods of time.
In \textbf{(a)}, we see a similar correlation between between the increase in the network's time-dependent accuracy (top row, left, in blue) and the modulus $|\lambda_\textrm{max}|$ of the largest eigenvalue of the fixed point (FP) associated with Symbol A (bottom row, left, in blue). \\
The learning process is more complex in \textbf{(b)}. The network's time-dependent accuracy plateaus around 0.5 initially and starts to rise from the plateau around training iteration 640. This escape from the plateau is short-lived, however, as the network quickly returns back to the plateau for training iterations 750 through 1200 before escaping a second time and converging to perfect accuracy. \\
The plot of $|\lambda_\textrm{max}|$ for Symbols A and B for the network in (b) also show a similar trend: the FPs are initially stable, become unstable temporarily when the network's TD accuracy spikes for the first time, but become stable again after the TD accuracy drops back to the plateau. This result further supports this correlation between the TD accuracy and the stability of the FPs of the network.\\
In (b), we also see another example of the null symbol having multiple fixed points that vanish before the end of the training.
}
\end{figure*}

\end{document}